\documentclass{Interspeech}

\interspeechcameraready

\title{Song Form-aware Full-Song Text-to-Lyrics Generation with \\ Multi-Level Granularity Syllable Count Control}

\author[affiliation={1}]{Yunkee}{Chae}
\author[affiliation={2}]{Eunsik}{Shin}
\author[affiliation={2}]{Suntae}{Hwang}
\author[affiliation={2}]{Seungryeol}{Paik}
\author[affiliation={1,2,3}]{Kyogu}{Lee}

\affiliation{Interdisciplinary Program in Artificial Intelligence}{Seoul National University}{}
\affiliation{Department of Intelligence and Information}{Seoul National University}{}
\affiliation{Artificial Intelligence Institute}{Seoul National University}{Republic of Korea}
\email{\{yunkimo95,esshin,iamsuntae1,paik402,kglee\}@snu.ac.kr}
\keywords{lyrics generation, syllable count control}

\usepackage{comment}

\usepackage{graphicx}

\usepackage{subcaption}

\usepackage{amsmath}
\usepackage{import}

\usepackage{booktabs} 
\usepackage{amssymb} 
\usepackage{pifont} 
\usepackage{makecell} 

\usepackage{multirow}
\usepackage{multicol}
\usepackage{array, hhline}
\usepackage{makecell}
\usepackage{subcaption}
\usepackage{caption}
\usepackage{arydshln}
\usepackage{lineno}
\usepackage{fix-cm}
\usepackage{hyperref}

\begin{document}

\maketitle
\begin{abstract}
Lyrics generation presents unique challenges, particularly in achieving precise syllable control while adhering to song form structures such as verses and choruses. 
Conventional line-by-line approaches often lead to unnatural phrasing, underscoring the need for more granular syllable management. 
We propose a framework for lyrics generation that enables multi-level syllable control at the word, phrase, line, and paragraph levels, aware of song form. 
Our approach generates complete lyrics conditioned on input text and song form, ensuring alignment with specified syllable constraints. 
Generated lyrics samples are available at: 
\url{https://tinyurl.com/lyrics9999}
\end{abstract}

\section{Introduction}

The field of lyrics information processing \cite{watanabe2020lip} presents unique challenges that extend beyond traditional text generation, as lyrics must align with both song form structures (such as verses, choruses, and bridges) and specific syllabic constraints. 
While natural language generation models have shown promise in generating coherent text, applying these models to lyrics introduces additional complexities due to the need for musical and rhythmic alignment.

Several studies have made notable progress in aligning lyrics with melodies through melody-aligned or midi-aligned generation methods \cite{watanabe2018melody, lu2019syllable, sheng2021songmass}. 
These efforts have been valuable in addressing melody-lyrics synchronization. 
However, the scarcity of large-scale audio-lyrics aligned datasets still poses a challenge for these approaches, limiting their generalizability across diverse musical forms \cite{meseguer2019dali, yu2021conditional, durand2023contrastive}.
Given this data limitation, we concentrate on a crucial aspect of melody-lyrics alignment: the syllable count, which directly impacts the rhythmic fit of lyrics within a musical composition.
However, precise syllable control is challenging, as models like GPT \cite{brown2020language}, which rely on sub-word tokens, may struggle to manage syllables effectively.

On the other hand, song form--the structured organization of lyrics into verses, choruses, and bridges--plays a crucial role in shaping both the narrative and thematic progression of songs. 
While some models generate lyrics effectively at a surface level, they often do not account for the intricate patterns imposed by song form, which is essential for producing musically coherent lyrics.

\begin{figure}[t!]
    \centering
    \includegraphics[width=0.8\columnwidth]{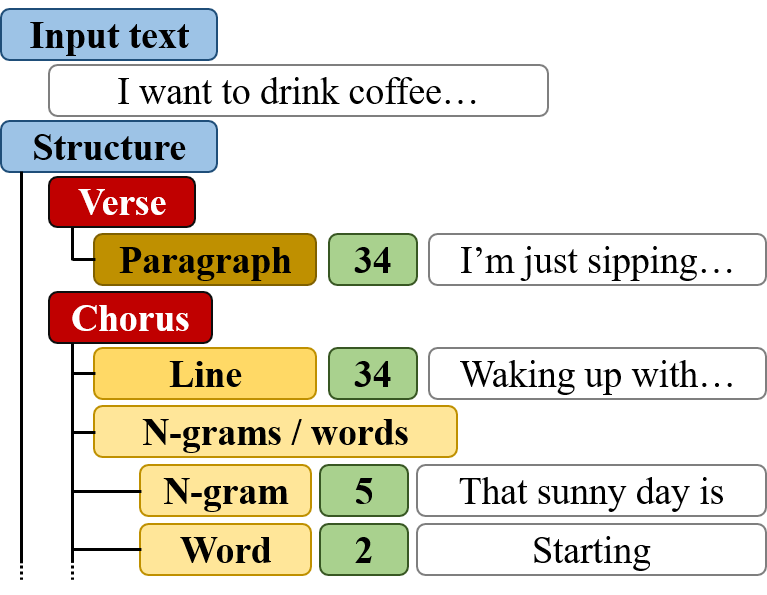}
    \vspace{-3mm}
    \caption{Multi-level granularity lyrics generation framework. The numbers inside the green boxes indicate the syllable count condition.}
    \label{fig:generation_framework}
    \vspace{-7mm}
\end{figure}

Therefore, we propose a comprehensive full-song lyrics generation system capable of operating under various conditions, including input text, song forms, and syllable counts across different granularities, as illustrated in Figure~\ref{fig:generation_framework}. 
Trained on a lyrics-only dataset, this system represents the first attempt, to the best of our knowledge, to address these challenges in an all-encompassing manner.

\section{Related Work}\label{sec:related_work}
Many studies on lyrics generation have attempted to use various features to generate lyrics that match naturally with the melody.
Lyrics generation inherits intricate syntactic and semantic challenges from text generation. Notably, within this domain, there has been a significant increase in the utilization of neural networks.

The procedure of converting melodies to expressive lyrics has also been studied, as seen in \cite{lu2019syllable,sheng2021songmass, chen2020melody, ma2021ai,  tian2023unsupervised}.
These models adeptly generate lyrics in coherence with the provided melody. 
For instance, \cite{chen2020melody} used SeqGANs \cite{yu2017seqgan} for this task, and \cite{ma2021ai} proposed AI Lyricists, leveraging MIDI files for lyrics generation.
Tailored specifically for melody-to-lyrics generation, models such as \cite{sheng2021songmass} utilize architectures like MASS \cite{song2019mass}.
Though these models address the problem in various ways, they suffer from a lack of melody-lyrics aligned dataset.
Approaches used in \cite{tian2023unsupervised} employ hierarchical frameworks successfully generating lyrics aligned with outlines, trained in an unsupervised manner to address this issue.
However, they do not consider song form or multi-granularity conditioned generation.

Researches focused on syllable count control has utilized rule-based and learning-based approaches to produce lyrics that adhere to constraints and convey meaning.
Poetry generation is a notable task in this field. 
For example, studies such as \cite{zhang2014chinese, yi2017generating, yi2018automatic} have employed RNN and LSTM for Chinese poetry generation. 
Additionally, \cite{lu2019syllable} proposed a method that interprets the alignment between lyrics and melodies as representations of syllable structures. This method uses a multi-channel sequence-to-sequence model that considers both phrasal structures and semantics.

\begin{figure}[t]
    \centering
    \begin{subfigure}[t]{.45\columnwidth}
        \centering
        \includegraphics[width=\textwidth]{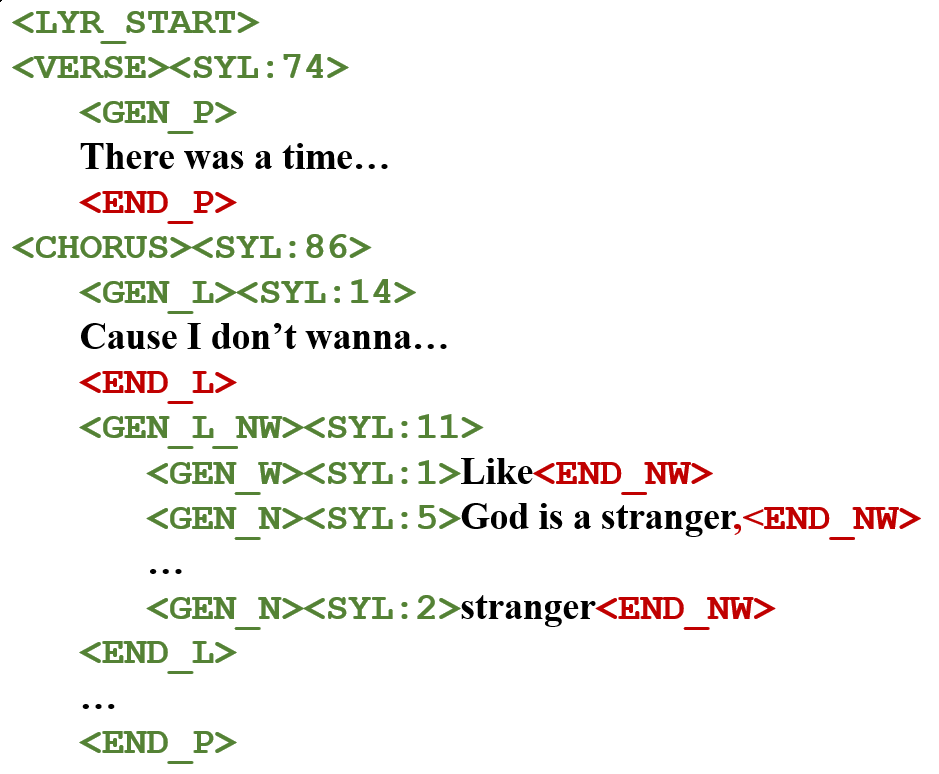}
        \caption{Generation sample}
        \label{subfig:train_gen}
    \end{subfigure}
    \begin{subfigure}[t]{.54\columnwidth}
        \centering
        \includegraphics[width=\textwidth]{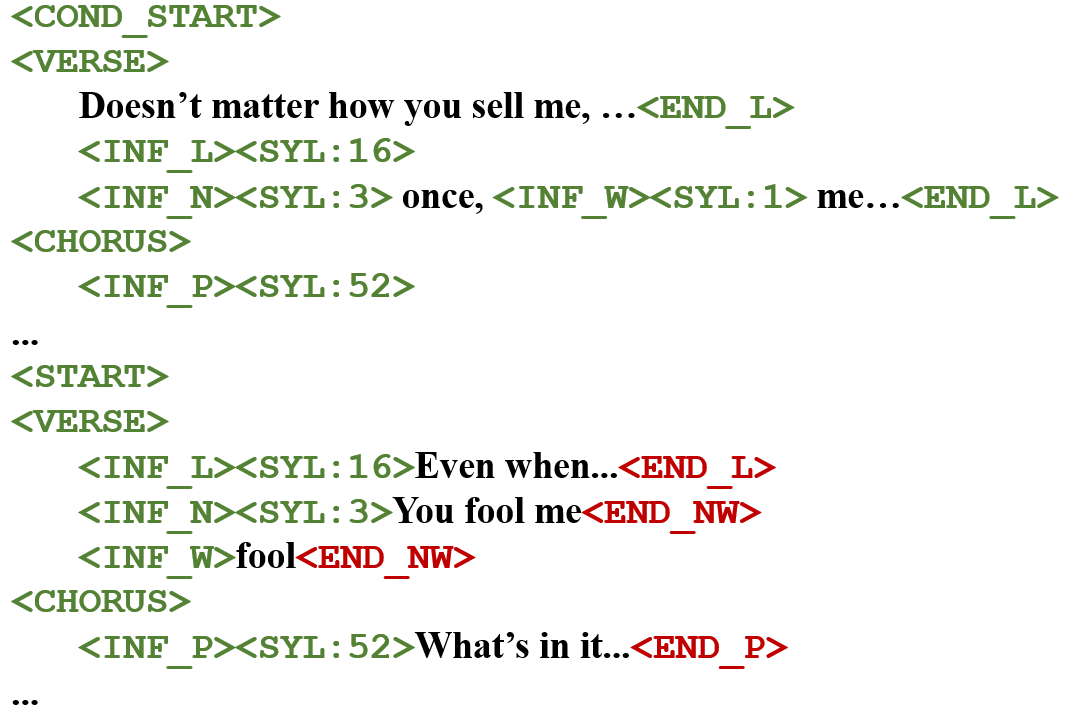}
        \caption{Infilling sample}
        \label{subfig:train_inf}
    \end{subfigure}
    \vspace{-2mm}
    \caption{Training examples of each task. The green tokens represent the tokens directly provided to the model during the inference process, while the red tokens denote the special tokens that the model needs to predict.}
    \label{fig:train_samples}
    \vspace{-6mm}
\end{figure}

\section{Methodology}\label{sec:method}
\subsection{Multi-Level Granularity Lyrics Generation}\label{subsec:lyrics_gen}
Our system generates lyrics at various granularities (words, phrases, lines, paragraphs) with precise syllable control. 
The model is conditioned on input text, song structure (e.g., verse, chorus), and syllable constraints.
\\
\textbf{Token Structure}.
We use a structured token system to align lyrics with song forms, as illustrated in Figure~\ref{subfig:train_gen}.
Each section begins with a song form token (e.g., \texttt{<VERSE>}) followed by a syllable count token (\texttt{<SYL:$s$>}), where $s$ denotes the designated syllable count for the section, adopting the approach described in \cite{kim2023k} and \cite{guo2022automatic}. 
Inspired by the ideas presented in \cite{bai-etal-2021-semantics}, we implement distinct end tokens for each level of granularity in our model.
For paragraphs, the generation starts with \texttt{<GEN\_P>} and ends with \texttt{<END\_P>}; 
entire lines use \texttt{<GEN\_L>} and \texttt{<END\_L>}. 
For finer segmentation, \texttt{<GEN\_L\_NW>} tokens are used to indicate the start of such detailed line generation.
Each segment, whether phrases or words, begins with \texttt{<GEN\_N>} or \texttt{<GEN\_W>} tokens, respectively, and concluded with \texttt{<END\_NW>} tokens.
Upon completing a line detailed with phrases and words, an \texttt{<END\_L>} token signifies its end. 
\\
\textbf{Semantic Embedding for Input Text}.
Our model also incorporates semantic embeddings for the input text, allowing it to generate contextually relevant lyrics.
Given the challenge of lacking datasets that pair text with lyrics, we devise an alternative training strategy that does not rely on such paired data.
We employ the \texttt{SentenceTransformer}
\cite{reimers-2019-sentence-bert} model to capture the semantic content of lyrics, a method inspired by the approach of Watanabe et al. \cite{watanabe2023text}, who used Stable Diffusion \cite{rombach2022high} and pre-trained Vision Transformer \cite{dosovitskiy2020image} for semantic extraction in lyrics.
We use semantic embeddings of full lyrics as a substitute condition for the input text during training, enabling the model to generate relevant lyrics by analyzing the semantic content of any input text provided during inference. 
Specifically, we incorporate the semantic embedding vector into the initial token embedding of the GPT-2 input.
\\
\textbf{Inference}.
During inference, the model starts with the semantic embedding from the input text and decodes tokens autoregressive. 
Special tokens such as song form, syllable counts, and generation directives tokens (\texttt{<GEN\_*>}) are provided at decoding step as conditions rather than predicted.
The model continues generating until it produces the appropriate end token (\texttt{<END\_*>}), then moves on to the next condition, following the complete generation plan.

\subsection{Lyrics Infilling}\label{subsec:lyrics_infill}
We also propose a lyrics infilling model that can adjust or correct lyrics based on multi-granularity syllable counts, drawing inspiration from \cite{donahue-etal-2020-enabling}.
As in Section \ref{subsec:lyrics_gen}, this method involves the introduction of special tokens designed to mask lyrics at various levels of granularity (e.g., \texttt{<INF\_P>}, \texttt{<INF\_L>}, etc.), as demonstrated in our infilling examples shown in Figure~\ref{subfig:train_inf}.
We randomly mask parts of the training data by replacing lyrics at different granularities with these tokens, along with their corresponding syllable count tokens.
The infilling process begins with the \texttt{<START>} token, followed by the song form tokens and the correct content for the masked segments, each marked with appropriate infilling and syllable tokens.

During inference, the model follows a similar procedure as Section \ref{subsec:lyrics_gen}, with the tokens preceding the \texttt{<START>} token serving as the contextual foundation and generates the missing parts.
Special tokens, including song form indicators, infilling directives (e.g. \texttt{<INF\_*>}), and syllable counts, are supplied directly to guide the generation, ensuring that each lyrical piece produced matches the specified conditions until the \texttt{<END\_*>} is reached.

\begin{table}[t!]
\centering
\footnotesize
\renewcommand{\arraystretch}{0.6}
\begin{tabular}{lccc}
\toprule
Models      & SCD $\downarrow$   & SCErr (\%) $\downarrow$ & BERT-S $\uparrow$\\
\hline
ChatGPT 3.5 & 0.363 & 83.729   & \textbf{0.897}\rule{0pt}{2.2ex}\\
ChatGPT 4   & 0.194 & 79.828   & 0.824  \\
\textit{Proposed}    & \textbf{0.004} & \textbf{4.396}   & 0.765 \\
\bottomrule
\end{tabular}
\caption{Comparison with large language models}
\label{tab:preliminary_llm_out}
\vspace{-10mm}
\end{table}

\begin{table*}[!t]
\begin{subtable}[h]{\textwidth}
\centering
\scriptsize
\fontsize{7.8}{8.2}\selectfont

\renewcommand{\arraystretch}{0.8}


\begin{tabular}
{l@{\hspace{2.5mm}}c@{\hspace{1mm}}c@{\hspace{1mm}}c@{\hspace{1mm}}c@{\hspace{1mm}}c@{\hspace{3mm}}c@{\hspace{1mm}}c@{\hspace{1mm}}c@{\hspace{1mm}}c@{\hspace{1mm}}c@{\hspace{3mm}}c@{\hspace{1mm}}c@{\hspace{1mm}}c@{\hspace{1mm}}c@{\hspace{1mm}}c@{\hspace{1.5mm}}c}

\toprule
\multirow{2}{*}{Model} & \multicolumn{5}{c}{\textbf{SCD} $\downarrow$} & \multicolumn{5}{c}{\textbf{SCErr} (\%) $\downarrow$} & \multicolumn{5}{c}{\textbf{PPL} $\downarrow$} & \multirow{2}{*}{\textbf{BERT-S} $\uparrow$} \\
\cmidrule(lr){2-6} \cmidrule(lr){7-11} \cmidrule(lr){12-16}
  & \textit{Full} & \textit{Para.} & \textit{Line} & \textit{Phrase} & \textit{Word} & \textit{Full} & \textit{Para.} & \textit{Line} & \textit{Phrase} & \textit{Word} & \textit{Full} & \textit{Para.} & \textit{Line} & \textit{Phrase} & \textit{Word} &  \\
 \hline
\textit{Front-P} & 0.026$^{*}$ & 0.068$^{*}$ & 0.013$^{*}$ & 0.021$^{*}$ & 0.024$^{*}$ & 10.313$^{*}$ & 84.856$^{*}$ & 11.711$^{*}$ & 6.140$^{*}$ & 3.618$^{*}$ 
& 17.069 & \textbf{31.259} & 28.442 & \textbf{16.793} & 11.271 & 0.735$^{*}$ \\

\textit{Front-S} & 0.006$^{*}$ & 0.047$^{*}$ & 0.004$^{*}$ & \textbf{0.002} & 0.002 & 5.501$^{*}$ & 75.781$^{*}$ & 3.998$^{*}$ & \textbf{0.759} & 0.252 
& 19.036 & 35.613 & 31.321 & 18.377 &12.218& 0.736$^{*}$ \\

\textit{Both-P} & 0.016$^{*}$ & 0.043$^{*}$ & 0.009$^{*}$ & 0.012$^{*}$ & 0.015$^{*}$ & 8.247$^{*}$ & 79.516$^{*}$ & 8.596$^{*}$ & 3.944$^{*}$ & 2.178$^{*}$ 
& \textbf{17.048} & 31.652 & \textbf{28.421} &16.831 & \textbf{11.259} & 0.737$^{*}$ \\

\textit{Both-S} & 0.005$^{*}$ & 0.041$^{*}$ & \textbf{0.003}$^{*}$ & \textbf{0.002} & \textbf{0.001} & 5.404$^{*}$ & 74.847$^{*}$ & 3.501$^{*}$ & 0.778 & \textbf{0.244} 
&  19.453 &  36.423 & 31.972 & 18.847 & 12.410 & 0.737$^{*}$ \\

\textit{Back-P} & 0.014$^{*}$ & 0.037$^{*}$ & 0.009$^{*}$ & 0.011$^{*}$ & 0.014$^{*}$ & 7.516$^{*}$ & 76.813$^{*}$ & 7.606$^{*}$ & 3.194$^{*}$ & 2.041$^{*}$ 
& 17.776 & 34.667 & 30.344 &  17.633 & 11.847 & 0.739$^{*}$ \\

 \hdashline
 \noalign{\vskip 1pt} 
\textit{Back-S} & \textbf{0.003} & \textbf{0.027} & \textbf{0.003} & \textbf{0.002} & 0.002 & \textbf{5.025} & \textbf{68.777} & \textbf{2.851} & 0.792 & 0.266 
& 19.883 & 39.189 & 33.867 & 19.422 & 12.802 & \textbf{0.740} \\
 \bottomrule

\end{tabular}

\caption{
Results of the generation task. Wilcoxon signed-rank tests are conducted between \textit{Back-S} and others.
The average BERT-S between the input text and the original lyrics is 0.799, which serves as an upper bound.
}
\label{subtab:result_gen}

\end{subtable}
\vspace{-2mm}
\begin{subtable}[h]{\textwidth}
\centering
\scriptsize
\fontsize{7.5}{8.0}\selectfont
\renewcommand{\arraystretch}{0.8}
\setlength{\tabcolsep}{0.6pt}
\begin{tabular}{lcccccccccc}

\toprule
\multirow{2}{*}{Model} & \multicolumn{5}{c}{\textbf{SCD $\downarrow$ / SCErr (\%)} $\downarrow$} & \multicolumn{5}{c}{\textbf{PPL $\downarrow$ / BERT-S $\uparrow$}} \\
\cmidrule(lr){2-6} \cmidrule(lr){7-11}
 & \textit{Full} & \textit{Para.} & \textit{Line} & \textit{Phrase} & \textit{Word} & \textit{Full} & \textit{Para.} & \textit{Line} & \textit{Phrase} & \textit{Word} \\
\hline
\noalign{\vskip 1pt} 
\multirow{1}{*}{\textit{LM-P}} 
& 0.024$^*$/5.226 & 0.037/75.763 & 0.015/7.186 & 0.009$^{*}$/2.542$^{*}$ & 0.020$^{*}$/1.493$^{*}$  & \textbf{7.425}/0.803$^{*}$ 
& 42.030$^{*}$/0.772$^{*}$  &\textbf{13.859}/0.746$^{*}$ & \textbf{5.371}/0.799$^{*}$ & \textbf{3.037}/0.839$^{*}$ \\

\multirow{1}{*}{\textit{LM-S}} 
& \textbf{0.002}/\textbf{3.327} & \textbf{0.034}/\textbf{72.860} & \textbf{0.004}/\textbf{4.576} & \textbf{0.001}/\textbf{0.593} & \textbf{0.001}/\textbf{0.092} & 7.887/0.812$^{*}$ & 48.166$^{*}$/0.781$^{*}$ & 14.591/0.754$^{*}$  & 5.596/0.807$^{*}$  & 3.087/0.851$^{*}$\\

\multirow{1}{*}{\textit{ILM-P}} 
& 0.008$^{*}$/6.642$^{*}$ & 0.065$^{*}$/82.151$^{*}$  & 0.016$^{*}$/15.685$^{*}$& 0.006$^{*}$/3.282$^{*}$ 
& 0.003$^{*}$/0.595$^{*}$& 11.675/0.855$^{*}$ &\textbf{21.021}/0.790$^{*}$& 21.388/0.785$^{*}$&12.516/0.848$^{*}$ 
& 6.913/\textbf{0.905} \\
\hdashline
\noalign{\vskip 1pt} 
\multirow{1}{*}{\textit{ILM-S}} & 0.006/5.479  & 0.047/75.140  & 0.013/11.744& 0.005/2.331 & 0.002/0.441 
&  12.897/\textbf{0.856} &  24.249/\textbf{0.812}  &  23.809/\textbf{0.798}& 13.829/\textbf{0.850}&  7.856/0.901 \\
\bottomrule
\end{tabular}

\caption{Results of the infilling task. Wilcoxon signed-rank tests are conducted between \textit{ILM-S} and others.}
\label{subtab:result_inf}
\end{subtable}

\flushleft
{
\vspace{-5mm}
\scriptsize
 Note: $^* p<0.005$
}
\caption{Results for generation and infilling task. We conducted one-tailed Wilcoxon signed-rank tests comparing each model to the one listed in the bottom row. The alternative hypotheses are that PPL, SCD, and SCErr are less, while the BERT-S hypothesis is greater.}
\label{tab:main_results}
\vspace{-8mm}
\end{table*}

\vspace{-2mm}
\section{Experiments}\label{sec:experiments}

\subsection{Dataset}
For the training and evaluation of our models, we used the 
Genius Song Lyrics dataset\cite{genius},
which comprises approximately 5.1 million multilingual song lyrics, with around 3.3 million lyrics in English. 
From the dataset, we selectively extracted lyrics featuring explicit song form annotations with each paragraph, aligning with a pre-defined set of song forms: verse, chorus, pre-chorus, post-chorus, and bridge.

To address the suitability of the content, we evaluated the toxicity of the lyrics using the \texttt{Detoxify} library \cite{Detoxify}, excluding any lyrics with a toxicity score greater than 0.5 to foster a positive and inclusive dataset.
Syllable counts for each word in the dataset were calculated using the \texttt{Syllables} library \cite{library_syllables}.

We split the dataset into approximately 340K training, 18K validation, and 10K evaluation samples.

\subsection{Preprocessing}
Due to the absence of a plan-lyrics paired dataset, we synthesize generation plans from lyric samples.
For each training example, we build a hierarchical tree (paragraphs, lines, words) and perform a pre-order traversal, randomly selecting subtrees with probability $p$, inspired by \cite{donahue-etal-2020-enabling}. 
For leaf nodes (word), we either include the word directly (50\% chance) or select a phrase (ranging from 1 to $min\{8, \text{\# of words remaining in the sentence}\}$). 
At the line level, all syllables are combined as shown in Figure~\ref{fig:train_samples}. 
This process informs the model of the required syllable counts in a line in advance.
For the infilling task, training samples are created similarly, but selected subtrees are replaced with mask tokens (i.e., \texttt{<INF\_*><SYL:$s$>}) and appended with the target text (see Section \ref{subsec:lyrics_infill} and Figure~\ref{subfig:train_inf}).
For the evaluation set, since there are no available input text-lyrics pairs, we created input-lyrics pairs by summarizing the lyrics with a summarization model
\cite{DBLP:journals/corr/abs-1910-13461}.

\begin{figure}[!t]
    \begin{subfigure}[t]{0.43\columnwidth}
        \centering
        \includegraphics[width=\columnwidth]{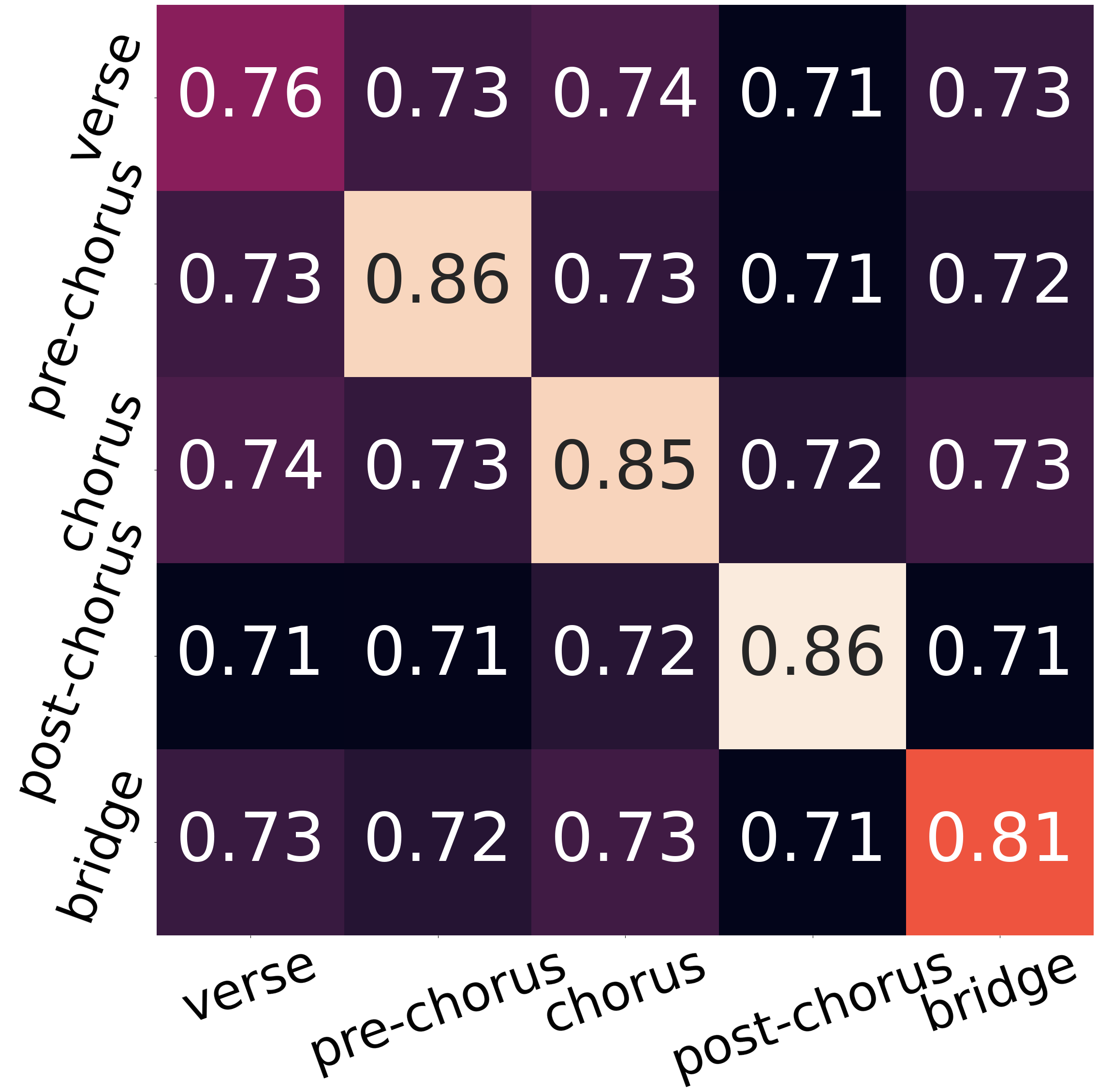}
        \vspace{-6mm}
        \caption{BERT-S}
        \label{subfig:song_form_bertscores}
    \end{subfigure}
    \begin{subfigure}[t]{0.43\columnwidth}
        \centering
        \includegraphics[width=\columnwidth]{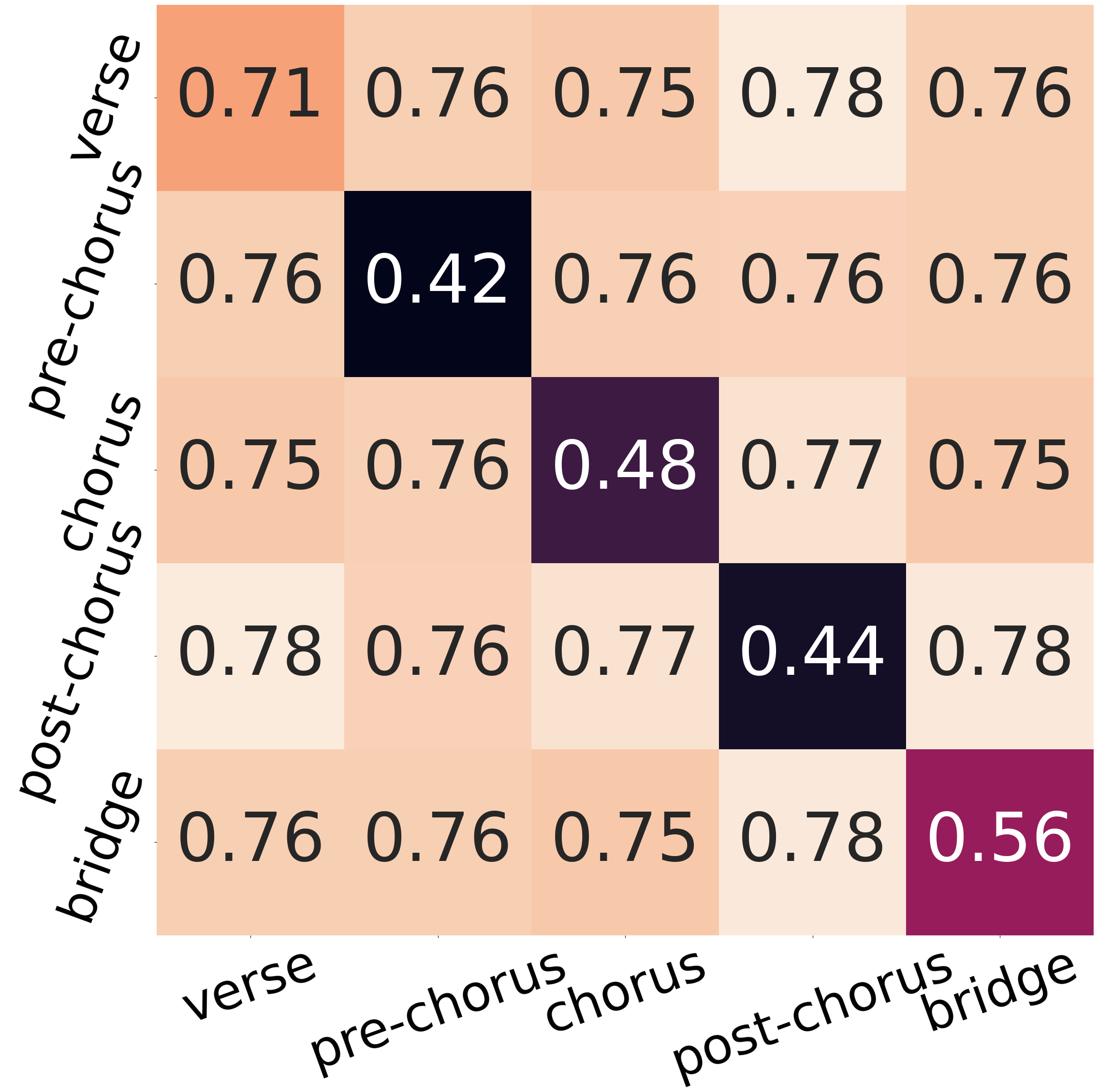}
        \vspace{-6mm}
        \caption{NLD}
        \label{subfig:song_form_levenshtein}
    \end{subfigure}
    \vspace{-3mm}
    \caption{Evaluation of song form consistency. Each metric is calculated between two distinct paragraphs.}
    \label{fig:songform_consistency}
    \vspace{-6mm}
\end{figure}

\subsection{Experimental Setup}
In all our experiments, we trained the GPT-2 model \cite{radford2019language_gpt} for 10 epochs, with the batch size of 8, using both scratch and pre-trained model.
The learning rate was 0.00005 with 500 warm-up steps, and we used the AdamW optimizer \cite{loshchilov2017decoupled}.
For all experiments, texts are tokenized using a byte-level Byte Pair Encoding (BPE) with a vocabulary size of 50,257 and an input size of 1024 consecutive tokens, as in the original GPT-2  model. 
For inference, we used top-k (20) and top-p (0.9) sampling with a temperature of 1.0 and a repetition penalty of 1.2. 
The pre-order tree traversal selection probability $p$ was set to 20\% for generation and 10\% for infilling.

\subsection{Evaluation Metrics}
In assessing the performance of our models, we employ a suite of metrics designed to capture various dimensions of model effectiveness, from linguistic predictability to syllabic accuracy and semantic fidelity.
\\
\textbf{Test set perplexity (PPL)} measures the ability of model to predict the text within the test set, providing insights into the model's linguistic prediction capabilities. 
PPL is computed on text tokens (excluding special tokens) and reported as a trimmed mean (top 1\% excluded). 
\\
\textbf{Syllable count distance (SCD)} \cite{kim2023computational} quantifies the discrepancy between the expected set of syllable counts $S=\{s_1,...,s_n\}$ and generated counts $\hat{S}=\{\hat{s}_1,...,\hat{s}_n\}$ as: 
\vspace{-2mm}
\begin{equation}
    SCD(S, \hat{S})=\frac{1}{2n}\sum_{i=1}^n({|s_i-\hat{s}_i|}/{s_i}+{|s_i-\hat{s}_i|}/{\hat{s}_i}).
    \label{eq:scd}
\end{equation}
\vspace{-4mm}
\\
\textbf{Syllable count error rate (SCErr)} measures how frequently the model generates text with incorrect syllable counts, representing a stricter assessment than SCD by penalizing all inaccuracies equally.
\\
\textbf{BERT-Score (BERT-S)} \cite{zhang2019bertscore} evaluates semantic coherence by comparing the input text with generated lyrics (or masked segments with infilled text) using pre-trained BERT embedding.
\\
\textbf{Normalized Levenshtein Distance (NLD)}
is also adopted to assess the consistency of the song form. 
Levenshtein distance, also known as edit distance, measures the number of character operations (deletions, insertions, or substitutions) required to convert one string into another. 
Normalized Levenshtein distance \cite{tashima2018fault} is defined as: 
\\
\vspace{-3mm}
\begin{equation}
    \text{NLD}(p_1, p_2) = {\text{LD}(p_1, p_2)}/{\max\{\lambda(p_1), \lambda(p_2)\}},
    \label{eq:nld}
\end{equation}
\vspace{-4mm}
\\
where $ \text{LD}(p_1, p_2) $ is the Levenshtein distance and $ \lambda(p) $ denotes the string length.
Metrics are computed overall (\textit{Full}) and at specific structural levels (e.g., paragraphs) for detailed insights.

\begin{table*}[!t]
\centering
\scriptsize
\fontsize{7.8}{8.2}\selectfont
\renewcommand{\arraystretch}{0.8}
\begin{tabular}{llllllllllllllll}
\toprule
 \multirow{2}{*}{Model} &\multicolumn{5}{c}{\textbf{SCD $\downarrow$ / SCErr (\%)} $\downarrow$} & \multicolumn{5}{c}{\textbf{PPL} $\downarrow$ / \textbf{BERT-S} $\uparrow$}  \\
 \cmidrule(lr){2-6} \cmidrule(lr){7-11}
& \textit{Full} & \textit{Para.} & \textit{Line} & \textit{Phrase} & \textit{Word} & \textit{Full} & \textit{Para.} & \textit{Line} & \textit{Phrase} & \textit{Word} \\
\hline
\multirow{2}{*}{\textit{ILM-S}} & \textbf{0.006} & \textbf{0.047} & 0.013 & \textbf{0.005} & \textbf{0.002} 
& \textbf{12.978} & \textbf{24.158} & \textbf{23.798} &  \textbf{13.979}  & \textbf{7.869} \rule{0pt}{2.2ex} \\
 & / 5.444 & / \textbf{74.952} & / 11.924 & / 2.405 & / 0.449 
&/ \textbf{0.856} &/ \textbf{0.813} & / \textbf{0.798}&/ \textbf{0.849} &/ \textbf{0.900} \\
\hdashline

\multirow{2}{*}{\textit{--same mask}} & 0.007 & 0.052$^{**}$ & 0.013 & \textbf{0.005} & 0.003$^{**}$ 
& 13.274$^{**}$& 24.277$^{**}$ &24.338$^{**}$ & 14.357$^{**}$ & 7.989$^{**}$ \rule{0pt}{2.2ex} \\
 & / 5.439 & / 77.601$^{**}$ & / 12.314$^{*}$ & / \textbf{2.240} & / 0.522$^{*}$ 
 &/ 0.853$^{**}$&/ \textbf{0.813} & / 0.795$^{**}$& / 0.845$^{**}$&/ 0.897$^{**}$ \\
 
\multirow{2}{*}{\textit{--no song form}} 
& \textbf{0.006} & 0.053$^{**}$ & \textbf{0.012} & \textbf{0.005} & \textbf{0.002} 
& 13.213$^{**}$ & 24.208$^{**}$ & 24.014$^{**}$ &  14.453$^{**}$ & 7.919$^{**}$ \\
 & / \textbf{5.318} & / 76.203$^{*}$ & / \textbf{11.771} & / 2.388 & / \textbf{0.046} 
 &/ 0.855$^{**}$ & / 0.812&/ \textbf{0.798} & / 0.848$^{**}$&/ 0.899$^{**}$ \\
 
\multirow{2}{*}{\textit{\begin{tabular}[c]{@{}l@{}}--same mask \&\\ no song form\end{tabular}}} 
& 0.007$^{**}$ & 0.054$^{**}$ & 0.014$^{**}$ & \textbf{0.005} & 0.003$^{**}$ 
& 13.623$^{**}$ &  24.514$^{**}$ & 24.146$^{**}$ & 14.836$^{**}$ & 8.588$^{**}$ \\
 & / 5.763$^{**}$ & / 79.741$^{**}$ & / 13.834$^{**}$ & / 2.449 & / 0.551$^{**}$ 
 &/ 0.852$^{**}$ &/ 0.812 & / 0.795$^{**}$& / 0.844$^{**}$& / 0.895$^{**}$\\
 
\bottomrule
\end{tabular}
\flushleft{
\vspace{-2mm}
\scriptsize
Note: $^* p<0.05, ^{**}p<0.005$}
\caption{Results of infilling variants. One-tailed Wilcoxon signed-rank tests are conducted to compare each model with \textit{ILM-S}.}
\label{tab:abl_inf_var}
\vspace{-8mm}
\end{table*}

\section{Results}
\subsection{Preliminary Experiment}\label{subsec:primary_experiments}
Large language models (LLMs) such as ChatGPT \cite{ouyang2022training} have demonstrated impressive text generation abilities but often struggle with accurate syllable counting \cite{sun-etal-2023-evaluating}.
To evaluate this, we asked an LLM to generate a single paragraph based on input text that specified the number of lines and the syllable count per line. 
A trial was deemed a failure if the model did not produce the correct number of lines within five attempts. 
ChatGPT 3.5 achieved a success rate of approximately 38\%, while ChatGPT 4.0 reached about 57\%. 
Table \ref{tab:preliminary_llm_out} summarizes the results for 688 successful samples, indicating that while LLMs are capable of generating lyrics that are contextually appropriate to the input text, they perform poorly in matching the required syllable count.

\subsection{Lyrics Generation}\label{subsec:result_gen}
In Table \ref{subtab:result_gen}, we present the performance of our models across various settings.
For each training method, we report on models trained from scratch as well as those initialized with large-scale pre-trained GPT-2 weights.

\textit{Front} models serve as our baseline. 
In these models, the complete generation plan is provided upfront—before the \texttt{<LYR\_START>} token—so that the model generates full lyrics without additional guidance during decoding. Essentially, we concatenate all special tokens into a single prompt to generate the output.
\textit{Back} models provide the generation plan to the model during decoding, as shown in Figure~\ref{subfig:train_gen}. 
\textit{Both} approach merges these methods: it first presents the complete generation plan before the \texttt{<LYR\_START>} token and then generates the lyrics using the \textit{Back} strategy. 
The suffixes \textit{-S} and \textit{-P} indicate whether the model was trained from scratch or initialized with pre-trained weights, respectively.

Table \ref{subtab:result_gen} shows that pre-trained models achieve lower PPL than models trained from scratch, with the \textit{Front} and \textit{Both} approaches performing best. 
A high PPL indicates that the model struggles to predict the correct tokens. 
The reason \textit{Front} and \textit{Both} models exhibit lower PPL compared to \textit{Back} models is likely because the model can preview all conditions before generating lyrics, enabling it to make more informed predictions.
In contrast, the \textit{Back} models receive conditioning at each decoding step, limiting their foresight and resulting in higher PPL because they cannot fully anticipate the complete structure.
However, given conditions directly at each decoding step could confer advantages in predicting syllable counts.
The \textit{Both} models attempt to integrate the strengths of both \textit{Front} and \textit{Back} models but at the expense of increased sample length.
Thus, each conditioning approach inherently involves a trade-off.

On the other hand, models trained from scratch consistently outperform pre-trained models in terms of syllable count accuracy, as measured by SCD and SCErr. 
This may be because scratch models learn to handle special tokens—like syllable tokens—more flexibly. 
Notably, the \textit{Back-S} model excels in both syllable metrics and BERT-S performance, indicating its superior ability to capture the input text’s semantics by generating immediately after the text embedding.

\subsection{Lyrics Infilling}\label{subsec:result_inf}
Table \ref{subtab:result_inf} presents the results for the infilling task.
For baseline comparisons, we use the same models as in the generation experiments—referred to as \textit{LM-S} and \textit{LM-P}—which generate masked segments using only past context (with unmasked tokens provided directly).
In contrast, our proposed Infilling Language Models (\textit{ILM-S} and \textit{ILM-P}) leverage both past and future context. 
While pre-trained models generally achieve lower PPL, models trained from scratch consistently yield better syllable accuracy.
\textit{LM} models exhibit lower PPL than \textit{ILM}s, and notably, \textit{LM-S} demonstrates superior syllable count accuracy.
These results suggest that leveraging comprehensive context does not necessarily improve syllable count accuracy.
However, \textit{ILM-S} attains higher BERT-S scores than its counterparts, indicating that using full context during infilling enhances semantic coherence. 
Overall, \textit{ILM-S} outperforms \textit{ILM-P} in both syllable metrics and BERT-S, suggesting that training from scratch better adapts to our specialized token structure. 
These findings are consistent with our observations from the lyrics generation experiments.

\subsection{Song Form Consistency Evaluation}
To verify that our model generates distinct lyrics for different song form types, we measure semantic similarity (BERT-S) and structural similarity (normalized Levenshtein Distance, NLD) between paragraphs. 
We compare only distinct paragraphs within the same song (e.g., verse1 vs. verse2, verse1 vs. chorus1) and compute the average values across all songs.

Figure~\ref{fig:songform_consistency} displays the results for the \textit{Back-S} model in a square matrix format. 
As these comparisons are made between paragraphs within the same song, we observe that the BERT-S generally exceeds 0.7. 
Notably, the BERT-S scores for comparisons within the same song form (diagonal values) are higher than those for comparisons between different forms.
Similarly, lower NLD values for same-form comparisons indicate that fewer edits are needed between similar segments.
Although verses exhibit slightly more variability due to their diverse content, they still show greater similarity within the same form compared to different forms. 
Overall, these results indicate that the model produces lyrics that are consistent within a given song form and distinct across different forms.
\subsection{Ablation Study}
Table~\ref{tab:abl_inf_var} examines the impact of modifications to our infilling strategy compared to the baseline \textit{ILM-S} model.
First, we train a variant using a uniform \texttt{<MASK>} token for all conditioning instead of specialized tokens (e.g., \texttt{<INF\_*><SYL:$s$>}) for different granularities; we denote this variant as \textit{same mask} in Table~\ref{tab:abl_inf_var}. 
In this setup, syllable condition tokens are provided directly at each infilling step. 
The results show a significant decline in syllable count performance at both the paragraph and word levels, along with reductions in overall BERT-S and increased PPL.
Next, we train a model without including song form tokens after the \texttt{<START>} token, denoted as \textit{no songform}. 
While this change has little effect on syllable count accuracy, it also results in a noticeable drop in BERT-S and a rise in PPL, similar to the \textit{same mask} condition.
Finally, we combine both modifications—using a uniform mask and omitting song form tokens—in a single model. 
This configuration degrades performance across nearly all metrics, confirming the importance of our specialized masking strategy and the inclusion of song form annotations for effective lyrics infilling.

\section{Conclusion}
Our study proposes an effective approach to automated lyrics generation and infilling that leverages multi-level granularity for precise syllable control and effective song form management, all conditioned on an arbitrary input text. This method enhances the flexibility and structural accuracy of the generated lyrics. In future work, we plan to incorporate additional controls—such as genre tags and rhyme schemes—to further improve the adaptability and creative potential of automated lyrics composition.

\section{Acknowledgements}
This work was supported by the National Research Foundation of Korea (NRF) grant funded by the Korea government (MSIT) [No. RS-2024-00461617, 40\%], Information \& communications Technology Planning \& Evaluation (IITP) grant funded by the Korea government(MSIT) [No. RS-2022-II220320, 2022-0-00320, 50\%], [No.RS-2021-II211343, Artificial Intelligence Graduate School Program (Seoul National University), 5\%], and [No.RS-2021-II212068, Artificial Intelligence Innovation Hub, 5\%]

\bibliographystyle{IEEEtran}
\bibliography{main}

\end{document}